\pdfoutput=1
 
\documentclass{article}

\usepackage{arxiv}

\usepackage[utf8]{inputenc} 
\usepackage[T1]{fontenc}    
\usepackage{hyperref}       
\usepackage{url}            
\usepackage{booktabs}       
\usepackage{amsfonts}       
\usepackage{nicefrac}       
\usepackage{microtype}      
\usepackage{lipsum}
\usepackage{graphicx}
\usepackage{color}
\usepackage{wrapfig}

\newcommand{\secref}[1]{(\textsection \ref{#1})}

\newcommand{\pddlgym}{\mbox{PDDLGym}}

\title{\pddlgym: Gym Environments from PDDL Problems}

\author{
  Tom Silver and Rohan Chitnis\\
  MIT Computer Science and Artificial Intelligence Laboratory\\
  \{tslvr, ronuchit\}@mit.edu
}
\begin{document}
\maketitle

\begin{abstract}
  We present \pddlgym{}, a framework that automatically constructs OpenAI Gym environments from PDDL domains and problems.
Observations and actions in \pddlgym{} are relational, making the framework particularly well-suited for research in relational reinforcement learning and relational sequential decision-making.
\pddlgym{} is also useful as a generic framework for rapidly building numerous, diverse benchmarks from a concise and familiar specification language.
We discuss design decisions and implementation details, and also illustrate empirical variations between the 20 built-in environments in terms of planning and model-learning difficulty.
We hope that \pddlgym{} will facilitate bridge-building between the reinforcement learning community (from which Gym emerged) and the AI planning community (which produced PDDL).
We look forward to gathering feedback from all those interested and expanding the set of available environments and features accordingly.

\end{abstract}

\section{Introduction}
\label{sec:intro}
The creation of benchmarks has often accelerated research progress in various subdomains of artificial intelligence~\cite{imagenet,glue,moleculenet}. 
In sequential decision-making tasks, tremendous progress has been catalyzed by benchmarks such as the environments in OpenAI Gym~\cite{openaigym} and the planning tasks in the International Planning Competition (IPC)~\cite{ipc}. 
Gym defines a standardized way for an agent to interact with an environment, allowing easy comparison of various reinforcement learning algorithms.
IPC provides a set of planning domains and problems written in the Planning Domain Definition Language (PDDL) \cite{pddl}, allowing easy comparison of various symbolic planners.

In this work, we present \pddlgym{}, an open-source framework that combines elements of Gym and PDDL. 
\textbf{\pddlgym{} is a Python library that automatically creates Gym environments \textit{from} PDDL domain and problem files.}

The library is available at \url{https://github.com/tomsilver/pddlgym}. Pull requests are welcome!

\begin{figure}[t]
  \centering
  \includegraphics[width=\textwidth]{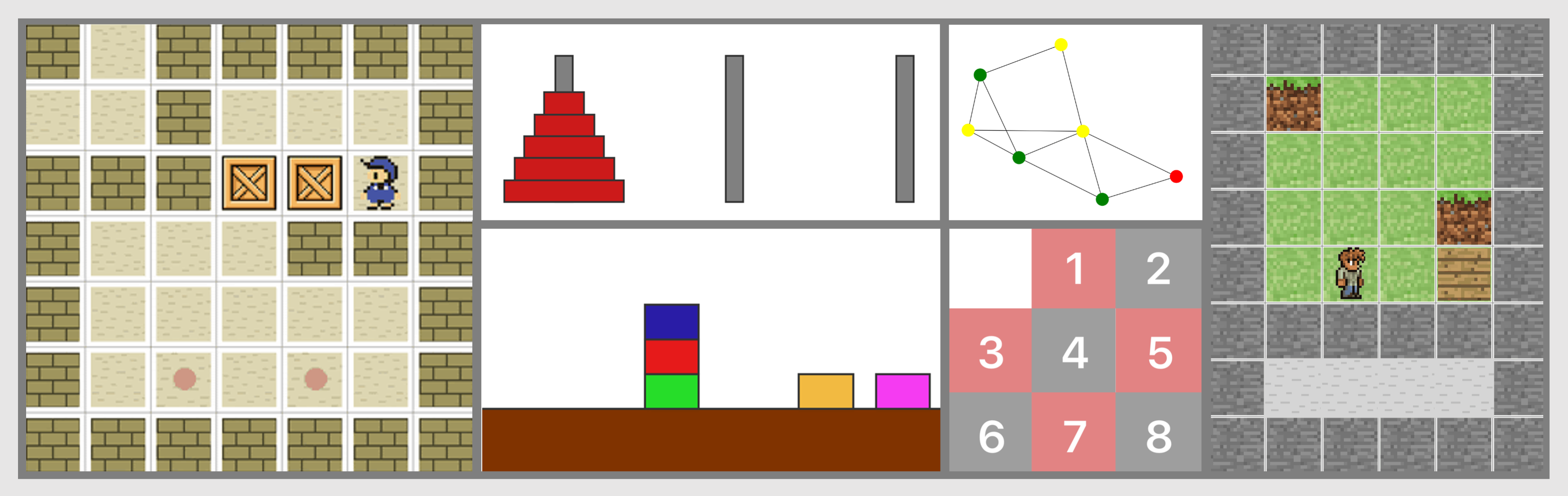}
  \caption{\textbf{Some examples of environments implemented in PDDLGym.} From top left: Sokoban, Hanoi, Blocks, Travelling Salesman (TSP), Slide Tile, and Crafting.}
  \label{fig:environments}
\end{figure}

\begin{figure}[t]
  \centering
  \includegraphics[width=\textwidth]{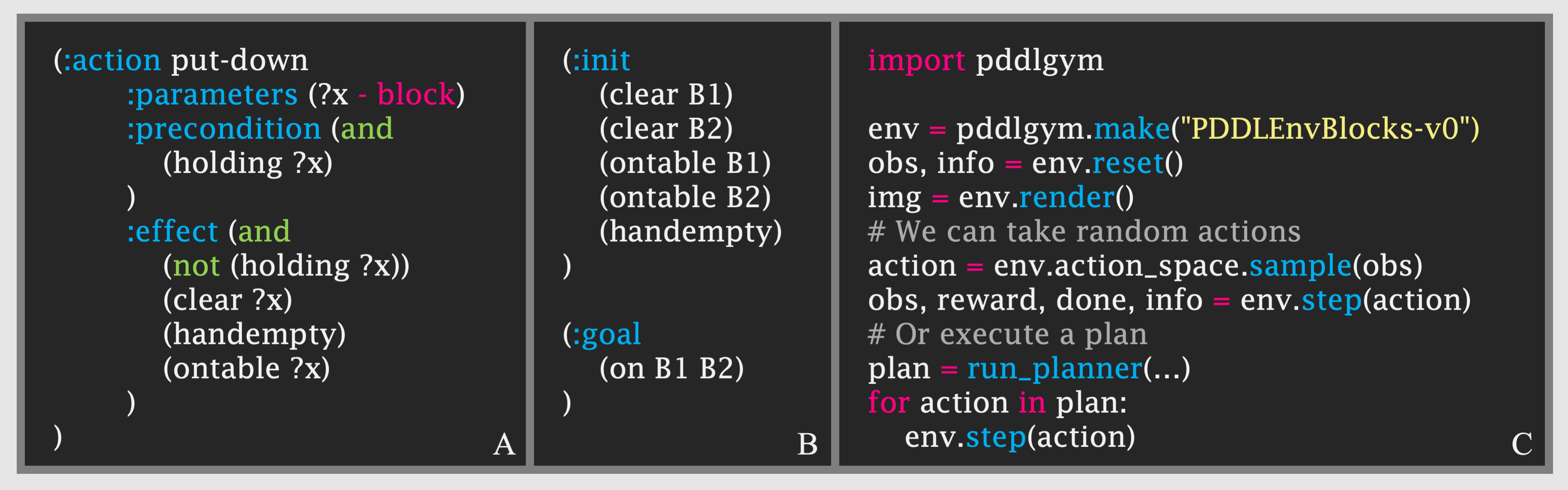}
  \caption{\textbf{\pddlgym{} code examples.} A \pddlgym{} environment is characterized by a PDDL domain file and a list of PDDL problem files. (A) One operator in the PDDL domain file for Blocks. (B) An excerpt of a simple PDDL problem file for Blocks. (C) After the PDDL domain and problem files have been used to register an environment with name ``PDDLEnvBlocks-v0,'' we can interact with this \pddlgym{} environment in just a few lines of Python.}
  \label{fig:example}
\end{figure}

As with Gym, \pddlgym{} allows for episodic, closed-loop interaction between the agent and the environment; the agent receives an observation from the environment and gives back an action, repeating this loop until the end of an episode.
As in PDDL, \pddlgym{} is fundamentally relational: observations are sets of ground relations over objects (e.g. \texttt{on(plate, table)}), and actions are templates ground with objects (e.g. \texttt{pick(plate)}).
\pddlgym{} is therefore particularly well-suited for relational learning and sequential decision-making research.
See Figure \ref{fig:environments} for renderings of some environments currently implemented in \pddlgym{}, and Figure \ref{fig:example} for code examples.

The Gym API used in reinforcement learning defines a hard boundary between the agent and the environment. 
In particular, the agent \emph{only} interacts with the environment by taking actions and receiving observations. 
The environment implements a function \texttt{step} that advances the state given an action by the agent; \texttt{step} defines the transition model of the environment. 
Likewise, a PDDL domain encodes a transition model via its operators. 
However, in typical usage, PDDL is understood to exist entirely in the ``mind'' of the agent. A separate process is then responsible for turning plans into actions that the agent can execute in the world.

\pddlgym{} defies this convention: in \pddlgym{}, PDDL domains and problems lie firmly on the environment side of the agent-environment boundary. The environment uses the PDDL files to implement the \texttt{step} function that advances the state given an action. \pddlgym{} is thus perhaps best understood as a \emph{repurposing} of PDDL.
Implementation-wise, this repurposing has subtle but important implications, discussed in \secref{sec:spaces}.

\newpage
\pddlgym{} serves three main purposes:

(1) \textit{Facilitate the creation of numerous, diverse benchmarks for sequential decision-making in relational domains}.
\pddlgym{} allows tasks to be defined in PDDL, automatically building a Gym environment from PDDL files.
PDDL offers a compact symbolic language for describing domains, which might otherwise be cumbersome and repetitive to define directly via the Gym API.

(2) \textit{Bridge reinforcement learning and planning research.}
\pddlgym{} makes it easy for planning researchers and machine learning researchers to test their methods on the exact same set of benchmarks, and to develop techniques that draw on the strengths of both families of approaches.
Furthermore, since \pddlgym{} includes built-in domains and problems, it is straightforward to perform apples-to-apples comparisons without having to collect third-party code from disparate sources (see also \cite{muise-icaps16demo-pd}).

(3) \textit{Catalyze research on sequential decision-making in relational domains.} 
In our own research, we have found \pddlgym{} to be very useful while studying exploration for lifted operator learning~\cite{glib}, hierarchical goal-conditioned policy learning~\cite{silver2020genplan}, and state abstraction~\cite{ploi}.
Other open research problems that may benefit from using \pddlgym{} include relational reinforcement learning~\cite{lang2012exploration,relational1,relational2}, learning symbolic descriptions of operators~\cite{lang2012exploration,amir2008learning,pasula2007learning}, discovering relational transition rules for efficient planning~\cite{xia2019learning,lang2010planning}, and learning lifted options~\cite{konidaris2014constructing,options1,options2,options3}.

The rest of this paper is organized as follows. 
\secref{sec:design} discusses the design decisions and implementation details underlying \pddlgym{}.
In \secref{sec:numbers}, we give an overview of the built-in \pddlgym{} domains and provide basic empirical results to illustrate their diversity in terms of the difficulty of planning and learning.
Finally, in \secref{sec:conclusion}, we discuss avenues for extending and improving \pddlgym{}.



\section{Design and Implementation}
\label{sec:design}
The Gym API defines environments as Python classes with three essential methods: \texttt{\_\_init\_\_}, which initializes the environment; \texttt{reset}, which starts a new episode and returns an observation; and \texttt{step}, which takes an action from the agent, advances the current state, and returns an observation, reward, a Boolean indicating whether the episode is complete, and optional debugging information.
The API also includes other minor methods, e.g., to handle rendering and random seeding.
Finally, Gym environments are required to implement an \texttt{action\_space}, which represents the space of possible actions, and an \texttt{observation\_space}, which represents the space of possible observations.
We next give a brief overview of PDDL files, and then we describe how action and observation spaces are defined in \pddlgym{}. Subsequently, we move to a discussion of our implementation of the three essential methods.
For implementation details regarding the main data structures used in \pddlgym{}, see \texttt{structs.py} in the code.

\subsection{Background: PDDL Domain and Problem Files}

There are two types of PDDL files: domain files and problem files.
A single \emph{benchmark} is characterized by one domain file and multiple problem files.

A PDDL domain file includes \textit{predicates} --- named relations with placeholder variables such as (\texttt{on~?x~?y}) --- and \textit{operators}.
An operator is composed of a name, a list of parameters, a first-order logic formula over the parameters describing the operator's preconditions, and a first-order logic formula over the parameters describing the operator's effects.
The forms of the precondition and effect formulas are typically restricted depending on the version of PDDL.
Early versions of PDDL only permit conjunctions of ground predicates \cite{strips}; later versions also allow disjunctions and quantifiers \cite{adl}.
See Figure \ref{fig:example}A for an example of a PDDL operator.

A PDDL problem file includes a set of \textit{objects} (named entities), an \textit{initial state}, and a \textit{goal}.
The initial state is a set of predicates ground with the objects.
Any ground predicates not in the state are assumed to be false, following the closed-world assumption.
The goal is a first-order logical formula over the objects (the form of the goal is limited by the PDDL version, like for operators' preconditions and effects).
Note that PDDL (and \pddlgym{}) also allows objects and variables to be \textit{typed}.
See Figure \ref{fig:example}B for a partial example of a PDDL problem file.

\subsection{Observation and Action Spaces}
\label{sec:spaces}


Each observation \texttt{obs} in \pddlgym{} has three components, mirroring the components of a PDDL problem file:
\texttt{obs.objects} is a set containing all objects present in the problem;
\texttt{obs.goal} contains the problem goal; 
and \texttt{obs.literals} is a set of all ground predicates that are true in the current state.
These observations fully encapsulate the state of the environment, i.e., \pddlgym{} environments are fully-observed.
The observation space is the powerset of all possible ground predicates, together with the objects and goal, which are static.
This powerset is typically enormous; fortunately, it usually does not need to be explicitly computed.
The observation space can also be viewed as a discrete space whose size is equal to the size of this powerset; since this space will be large, we expect that most algorithms for solving \pddlgym{} tasks will not be sensitive to its size.

The action space for a \pddlgym{} environment is one of the more subtle aspects of the overall framework, and there are two possible avenues to take. Instructions for taking both avenues are provided in the repository's README, in the ``Step 3: Register Gym environment'' section. 

The first avenue is appropriate if one wants to simply use off-the-shelf PDDL files with PDDLGym. One can do so by setting \texttt{operators\_as\_actions} to \texttt{True} in the environment registration, which tells PDDLGym that the operators present in the PDDL domain file should themselves be treated as the actions in the environment, parameterized by those operators' parameters.

\begin{figure}[t]
  \centering
  \includegraphics[width=\textwidth]{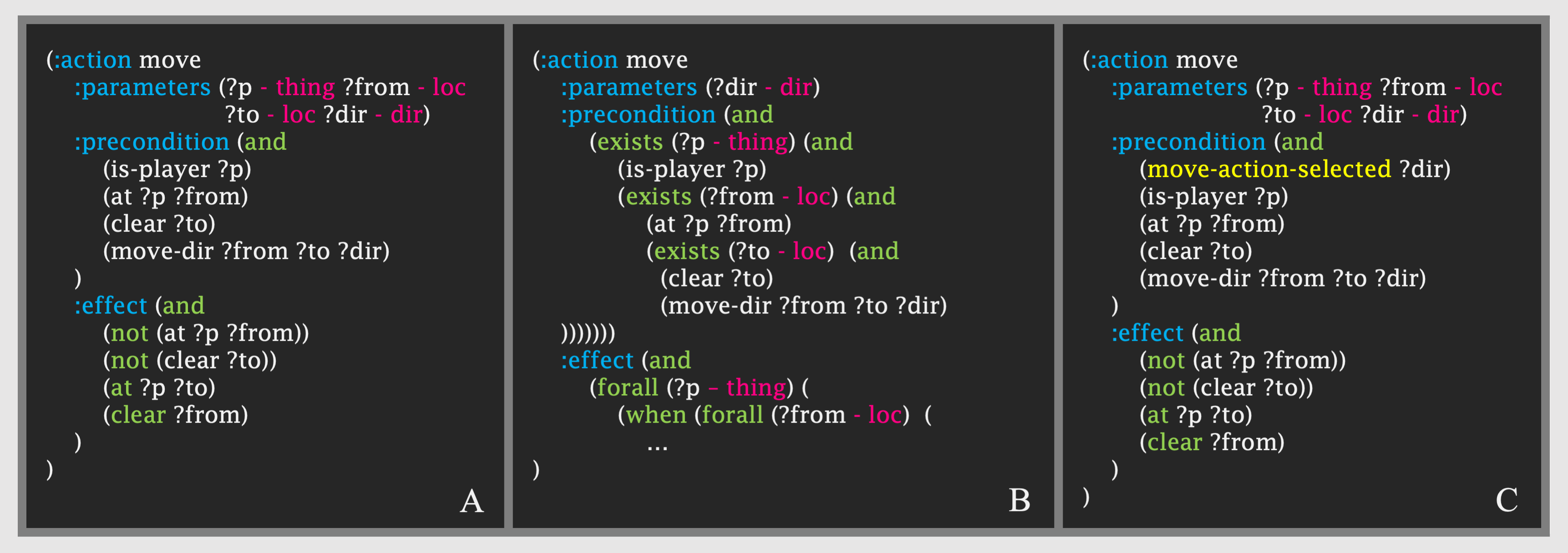}
  \caption{\textbf{Explicating free parameters in PDDL operators.} PDDL operators traditionally conflate free and non-free parameters. For example, in a typical move operator for Sokoban (A), the free parameter \texttt{?dir} is included alongside non-free parameters. \pddlgym{} must distinguish free parameters to properly define the action space. One option would be to require that all operator parameters are free, and introduce quantifiers in the operator body accordingly (B); however, this is cumbersome and leads to clunky, deeply nested operators, so we do \emph{not} do this. Instead, we opt to introduce new predicates that are tied to operators, and whose parameters are just the operators' free parameters (C). An example of such a new predicate is shown in yellow (\texttt{move-action-selected}).}
  \label{fig:actioncode}
\end{figure}

The second avenue is recommended for more serious research, and stems from the semantic difference between ``operators'' in classical AI planning and ``actions'' in reinforcement learning.
In AI planning, actions are typically equated with ground operators --- operators whose parameters are bound to objects.
However, in most PDDL domains, only some operator parameters are \textit{free} (in terms of controlling the agent); the remaining parameters are included in the operator because they are part of the precondition/effect expressions, but can be derived from the current state or the choice of free parameters.
PDDL itself makes no distinction between free and non-free parameters.
For example, consider the operator for Sokoban shown in Figure \ref{fig:actioncode}A. This operator represents the rules for a player (\texttt{?p}) moving in some direction (\texttt{?dir}) from one cell (\texttt{?from}) to another cell (\texttt{?to}).
In a real game of Sokoban, the only choice that an agent makes is what direction to move --- only the \texttt{?dir} parameter is free. The player \texttt{?p} is always the same, \texttt{?from} is defined by the agent's location in the current state, and \texttt{?to} can be derived from \texttt{?from} and the agent's choice of \texttt{?dir}.
To properly define the action space for a \pddlgym{} environment, we must explicitly distinguish free parameters from non-free ones.
One option is to require that operator parameters are all free.
Non-free parameters could then be folded into the preconditions and effects using quantifiers \cite{adl}; see Figure \ref{fig:actioncode}B for an example.
However, this is cumbersome and leads to clunky, deeply nested operators.
Instead, we opt to introduce new predicates that represent operators, and whose variables are these operators' free parameters. We then include these predicates in the preconditions of the respective operators; see Figure \ref{fig:actioncode}C for an example.
Doing so requires only minimal changes to existing PDDL files and does not affect readability, but requires adding in domain knowledge about the agent-environment boundary. Note that this domain knowledge is equivalent to defining an action space, which is very commonly done in reinforcement learning and is not a strong assumption. In this case, the action space of a \pddlgym{} environment is a discrete space over all possible groundings of the newly introduced predicates.

When sampling from the action space of a \pddlgym{} environment, \pddlgym{} will automatically only sample \emph{valid} actions, i.e., actions that satisfy the preconditions of some operator. This check for validity is done using Fast Downward's translator~\cite{fd}, which can add non-negligible overhead in large problem instances.

\subsection{Initializing and Resetting an Environment}

A \pddlgym{} environment is parameterized by a PDDL domain file and a list of PDDL problem files.
For research convenience, each \pddlgym{} environment is associated with a \emph{test} version of that environment, where the domain file is identical but the problem files are different (for instance, they could encode more complicated planning tasks, to measure generalizability).
During environment initialization, all of the PDDL files are parsed into Python objects; we use a custom PDDL parser for this purpose.
When \texttt{reset} is called, a single problem instance is randomly selected.\footnote{Problem selection when resetting an episode is the only use of randomness in \pddlgym{} (aside from stochastic transitions).}
The initial state of that problem instance is the state of the environment.
For convenience, \texttt{reset} also returns (in the debugging information) paths to the PDDL domain and problem file of the current episode.
This makes it easy for a user to run to a symbolic planner and execute resulting plans in the environment; see the README in the \pddlgym{} \href{https://github.com/tomsilver/pddlgym}{Github repository} for an example that uses Fast-Forward \cite{ff}.

\subsection{Implementing \texttt{step}}

The \texttt{step} method of a \pddlgym{} environment takes in an action, updates the environment state, and returns an observation, reward, done Boolean, and debugging information.
To determine the state update, \pddlgym{} checks whether any PDDL operator's preconditions are satisfied given the current state and action.
Note that it is impossible to ``accidentally match'' to an undesired operator: each operator has a unique precondition as illustrated in Figure~\ref{fig:actioncode}C, which is generated automatically based on the passed-in action.
Since actions are distinct from operators \secref{sec:spaces}, this precondition satisfaction check is nontrivial; non-free parameters must be bound. 
We have implemented two inference back-ends to perform this check.
The first is a Python implementation of typed SLD resolution, which is the default choice when the query involves only conjunctions.
The second is a wrapper around SWI Prolog \cite{swiprolog}, which permits us to handle more sophisticated preconditions involving disjunctions and quantifiers.
The latter is slower, but more general, than the former.
When no operator preconditions hold for a given action, the state remains unchanged by default.
In some applications, it may be preferable to raise an error if no preconditions hold; the optional initialization parameter \texttt{raise\_error\_on\_invalid\_action} permits this behavior.

Rewards in \pddlgym{} are sparse and binary.
In particular, the reward is 1.0 when the problem goal is satisfied and 0.0 otherwise.
Similarly, the done Boolean is True when the goal is reached and False otherwise.
(In practice, a maximum episode length is often used.)

If the underlying PDDL domain has probabilistic effects, as in PPDDL~\cite{ipc2008}, the \texttt{step} method will parse this appropriately and choose an effect based on the given probability distribution. If the given probabilities do not sum to 1, a default trivial effect is added in.

\subsection{Development Status}

In terms of lines of code, the bulk of \pddlgym{} is dedicated to PDDL file parsing and inference (used in \texttt{step}).
We are continuing to develop both of these features so that a wider range of PDDL domains are supported.
Aspects of PDDL 1.2 that are supported by \pddlgym{} include STRIPS, hierarchical typing, equality, quantifiers, constants, and derived predicates. Notable features that are not supported include conditional effects and action costs.
Aspects of later PDDL versions, such as numerical fluents, are not supported.
Our short-term objective is to provide full support for PDDL 1.2.
We have found that a wide range of standard PDDL domains are already well-supported by \pddlgym{}; see \secref{sec:numbers} for an overview. We welcome requests for features and extensions, via either issues created on the Github page or email. The authors' email addresses are provided at the top of this document.

\section{\pddlgym{} by the Numbers}
\label{sec:numbers}
In this section, we start with an overview of the domains built into \pddlgym{}, as of the last date this report was updated (\today).
We then provide some experimental results that give insight into the variation between these domains, in terms of planning and model-learning difficulty.
All experiments are performed on a single laptop with 32GB RAM and a 2.9GHz Intel Core i9 processor.

\subsection{Overview of Environments}

\begin{table}[t]
  \centering
  \resizebox{0.6\columnwidth}{!}{
  \tabcolsep=0.3cm{
  \begin{tabular}{ccccc}
    \toprule[1.5pt]
     \textbf{Domain Name} & \textbf{Source} & \textbf{Rendering Included} & \textbf{Probabilistic} & \textbf{Average FPS}\\
    \midrule[2pt]
    \textbf{Baking} & Ours & No & No & 5897 \\
    \midrule
    \textbf{Blocks} & \cite{pyperplan} & Yes & No & 7064 \\
    \midrule
    \textbf{Casino} & Ours & No & No & 7747 \\
    \midrule
    \textbf{Crafting} & Ours & Yes & No & 4568 \\
    \midrule
    \textbf{Depot} & \cite{pyperplan} & No & No & \hspace{1em}97 \\
    \midrule
    \textbf{Doors} & \cite{lightworld} & Yes & No & \hspace{0.5em}917 \\
    \midrule
    \textbf{Elevator} & \cite{pyperplan} & No & No & 3501 \\
    \midrule
    \textbf{Exploding Blocks} & \cite{ipc2008} & Yes & Yes & 6260 \\
    \midrule
    \textbf{Ferry} & \cite{csupddl} & No & No & 1679 \\
    \midrule
    \textbf{Gripper} & \cite{soarpddl} & Yes & No & \hspace{0.5em}319 \\
    \midrule
    \textbf{Hanoi} & \cite{soarpddl} & Yes & No & 4580 \\
    \midrule
    \textbf{Meet-Pass} & \cite{csupddl} & No & No & 7380
    \\
    \midrule
    \textbf{Rearrangement} & Ours & Yes & No & 3808
    \\
    \midrule
    \textbf{River} & \cite{ipc2008} & No & Yes & \hspace{-0.5em}18632 \\
    \midrule
    \textbf{Search and Rescue} & Ours & Yes & No & 3223 \\
    \midrule
    \textbf{Slide Tile} & \cite{soarpddl} & Yes & No & 3401 \\
    \midrule
    \textbf{Sokoban} & \cite{pyperplan} & Yes & No & \hspace{0.5em}155 \\
    \midrule
    \textbf{Triangle Tireworld} & \cite{ipc2008} & No & Yes & 6491 \\
    \midrule
    \textbf{TSP} & \cite{soarpddl} & Yes & No & 1688 \\
    \midrule
    \textbf{USA Travel} & Ours & No & No & 1251 \\
    \bottomrule[1.5pt]
  \end{tabular}}}
    \vspace{8pt}
  \caption{\textbf{List of the 20 domains currently included in \pddlgym{}, as of the last date this report was updated (\today).} For each environment, we report the original source of the PDDL files, whether or not we have implemented custom rendering, whether or not the domain has probabilistic effects, and the average frames per second (FPS). The FPS is calculated by executing a random policy for 100 episodes of 10 timesteps each, with no rendering.}
  \label{tab:environments}
\end{table}

There are currently 20 domains built into \pddlgym{}.
Most of the domains are adapted from existing PDDL repositories; the remainder are ones we found to be useful benchmarks in our own research.
We have implemented custom rendering for 11 of the domains (see Figure \ref{fig:environments} for examples). Table \ref{tab:environments} gives a list of all environments, their sources, and their average frames per second (FPS) calculated by executing a random policy for 100 episodes of 10 timesteps each, with no rendering.

\subsection{Variation in Environment Difficulty}

\begin{figure}[t]
  \centering
  \includegraphics[width=\textwidth]{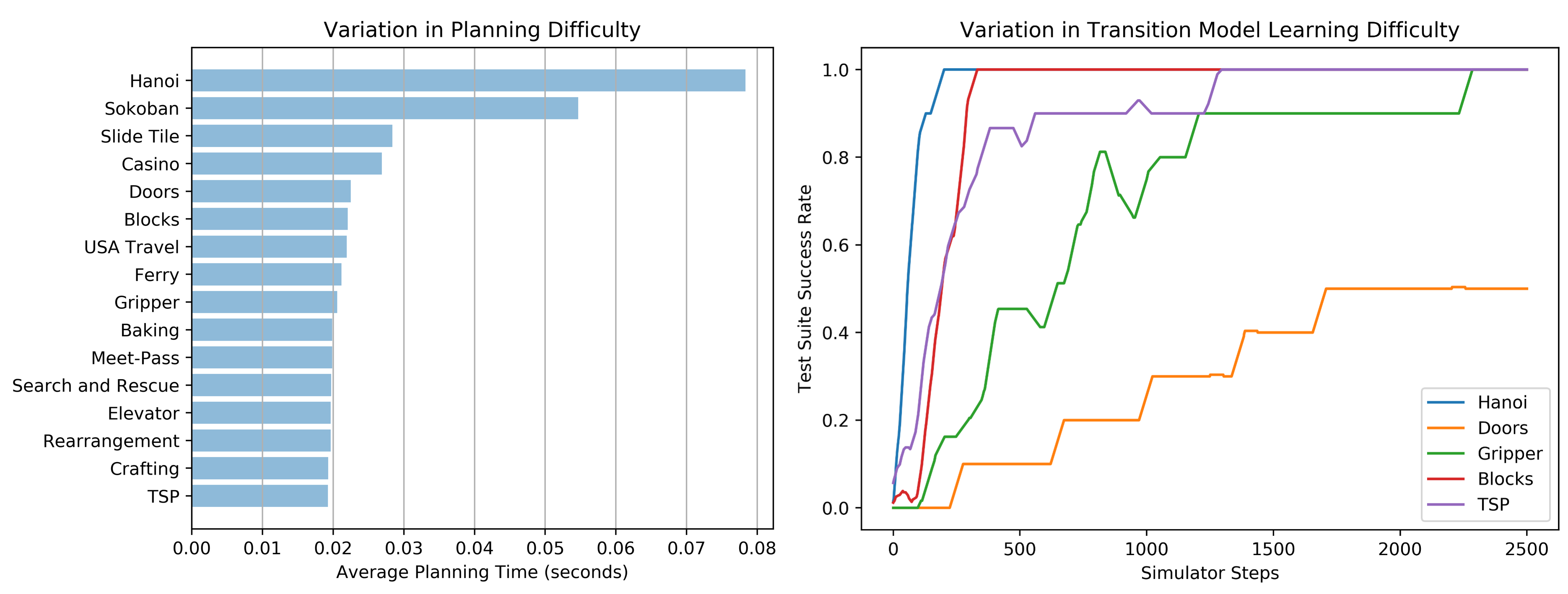}
  \caption{\textbf{Variation among \pddlgym{} environments.} The PDDL domains and problems built into \pddlgym{} vary considerably in terms of planning difficulty (left) and model learning difficulty (right). See text for details. One of the domains, Depot, was omitted for visual clarity, but required two orders of magnitude more planning time than the simplest one (TSP).}
  \label{fig:plots}
\end{figure}

We now provide some results illustrating the variation between the domains built into \pddlgym{}.
We examine two axes of variation: planning difficulty and difficulty of learning the transition model.

Figure \ref{fig:plots} (left) illustrates the average time taken by Fast-Forward \cite{ff} to find a plan in each of the deterministic environments, averaged across all problem instances.
The results reveal a considerable range in planning time, with the most difficult domain (Depot, omitted from the plot for visual clarity) requiring two orders of magnitude more time than the simplest one (TSP).
The results also indicate that many included domains are relatively ``easy'' from a modern planning perspective. However, even in these simple domains, there are many interesting challenges to be tackled, such as learning the true PDDL operators from interaction data, or defining good state abstractions amenable to learning. One can always make larger problem instances if desired, to push the limits of modern planners.

Figure \ref{fig:plots} (right) provides insight into the difficulty of learning transition models in some of the environments.
For each environment, an agent executes a random policy for episodes of horizon 25.
The observed transitions are used to learn transition models, which are then used for planning on a suite of test problems.
The fraction of test problems solved is reported as an indicator of the learned transition model.
To learn the transition models, we use first-order logic decision tree (FOLDT) learning \cite{foldt}.
Five domains are visualized for clarity;
among the remaining ones, several are comparable to the ones shown, but others, including Baking, Depot, and Sokoban, are difficult for our learning method: FOLDT learning is unable to find a model that fits the data in a reasonable amount of time.
Of course, model-learning difficulty varies considerably with the learning method and the exploration strategy. We have implemented simple strategies here to show these results, but these avenues for future research are exactly the kind that we hope to enable with \pddlgym{}.


\section{Conclusion and Future Work}
\label{sec:conclusion}
We have presented \pddlgym{}, an open-source Python framework that automatically creates OpenAI Gym environments from PDDL domain and problem files.
Our empirical results demonstrate considerable diversity among the built-in environments.
We have been using \pddlgym{} actively in our own research on relational sequential decision-making and reinforcement learning.
We also hope to interface \pddlgym{} with other related open-source frameworks, particularly the PDDL collection and tools in \url{planning.domains} \cite{muise-icaps16demo-pd}, so that a user can use \pddlgym{} simply by specifying a URL pointing to a repository of PDDL files (along with some domain-specific information about free parameters).

We look forward to gathering feedback from the community and expanding the set of available environments and features accordingly.

\bibliographystyle{unsrt}  
\bibliography{references}

\end{document}